\documentclass[letterpaper, 10 pt, conference]{ieeeconf}  

\usepackage{algorithm}
\usepackage{algpseudocode}
\usepackage{dirtytalk}
\usepackage{graphicx} 
\usepackage{booktabs}
\usepackage{tabularx}
\usepackage{hyperref}

\IEEEoverridecommandlockouts                              

\usepackage{graphics} 
\usepackage{epsfig} 
\usepackage{mathptmx} 
\usepackage{times} 
\usepackage{amsmath} 
\usepackage{amssymb}  

\title{\LARGE \bf
Selective Exploration and Information Gathering in Search and Rescue Using Hierarchical Learning Guided by Natural Language Input}

\author{Dimitrios Panagopoulos$^{1}$, Adolfo Perrusqu\'{i}a$^{1}$ and Weisi Guo$^{1}$
\thanks{$^{1}$School of Aerospace, Transport and Manufacturing, Cranfield University, Bedford, UK,
        {\tt\small \{d.panagopoulos, adolfo.perrusquia-guzman, weisi.guo\}@cranfield.ac.uk}}%
}

\begin{document}
\bstctlcite{IEEEexample:BSTcontrol}

\maketitle
\thispagestyle{empty}
\pagestyle{empty}

\begin{abstract}
In recent years, robots and autonomous systems have become increasingly integral to our daily lives, offering solutions to complex problems across various domains. Their application in search and rescue (SAR) operations, however, presents unique challenges. Comprehensively exploring the disaster-stricken area is often infeasible due to the vastness of the terrain, transformed environment, and the time constraints involved. Traditional robotic systems typically operate on predefined search patterns and lack the ability to incorporate and exploit ground truths provided by human stakeholders, which can be the key to speeding up the learning process and enhancing triage. Addressing this gap, we introduce a system that integrates social interaction via large language models (LLMs) with a hierarchical reinforcement learning (HRL) framework. The proposed system is designed to translate verbal inputs from human stakeholders into actionable RL insights and adjust its search strategy. By leveraging human-provided information through LLMs and structuring task execution through HRL, our approach not only bridges the gap between autonomous capabilities and human intelligence but also significantly improves the agent's learning efficiency and decision-making process in environments characterised by long horizons and sparse rewards.
\end{abstract}

\section{INTRODUCTION} \label{introduction}
Autonomous intelligent robots are expected to be deployed in a broader range of real-world applications \cite{SOORI202354}. For instance, in the aftermath of natural or man-made disasters, search and rescue (SAR) robots are utilised to assist in tasks such as searching operations, incidents reporting, locating missing people, and providing aid to those affected within the impacted area \cite{el2024advanced}. In this high-stakes domain, despite considerable advancements, SAR robots continue to face significant limitations in terms of decision-making, task execution, and adaptability. These limitations stem from the robot's dependency on preset behaviors, significant environment change/degradation, and accurate data input by human operators (e.g., upload updated viable roads/bridges \cite{bridge, damage} and triage priority areas for SAR \cite{ liu2013robotic}). However, this is in contention with the future vision of having fully-autonomous agents with the capacity to emulate human decision-making processes and adapting in real-time \cite{xi2023rise}, especially within human-robot interaction (HRI) regimes. 

\subsection{State-of-the-Art and Gaps}
Here, we consider how to make use of the user contributed information provided into the learning scheme. The literature suggests various forms of human guidance to aid learning algorithms, ranging from demonstrations and advice to preference and online evaluative feedback \cite{zhang2019leveraging}. The proposed work, nevertheless, is specifically designed for scenarios that humans have no access to the agent's internal software implementation. Instead, the feedback in these instances is communicated through natural language and converted into grounded insight. In fact, we know information theoretical bounds to information aided human-robot SAR exist \cite{human-SAR}, but how this can be achieved in reality is not clear. 

In this research area, there is widespread acknowledgement that the utility of informative feedback is beneficial to the learning agent \cite{bignold2022human}. Information about the environment's structure is densely informative, helping the agent reduce exploration and quickly find the optimal strategy \cite{guo2022survey}. Early work examined the use of side information from mobile signals or sensor networks \cite{phone}, however this lacks semantic detail and availability.

Current SAR systems deployed in these scenarios fail to actively seek, collect, and exploit contextual information from human stakeholders, which can be important to speeding up the learning progress and enhancing the efficiency of the search. This oversight results in a significant under-utilisation of potentially vital information and insights that could otherwise direct efforts more accurately and promptly \cite{10.1371/journal.pone.0273696}. In contrast, humans naturally tend to seek assistance when faced with challenging tasks, particularly when there is a lack of sufficient information and knowledge regarding the operational environment. This inclination to request help originates from our inherent desire for collaboration, problem-solving, and leveraging collective knowledge and expertise \cite{Kozlowski2006Teams}.

In view of the above, the challenge lies in extending SAR robots' capabilities beyond mere execution of tasks to becoming active participants in the problem solving process. This is especially crucial in large-scale and dynamic disaster environments, where much of the necessary information, initially unknown or inaccessible to rescue personnel, is scattered across various sources \cite{waring2020role}. Recognising the dynamic nature of such environments, where prior knowledge may be limited or outdated, we propose the integration of human linguistic inputs into the agent's learning process as a critical enhancement over reliance solely on environmental cues.

\subsection{Opportunities in LLMs and HRL}
Large Language Models (LLMs) \cite{LLMsEvaluation}, with their advanced natural language understanding capabilities, emerge in bridging the communication gap between SAR robots and humans by providing a variety of constructive roles in solving planning tasks \cite{kambhampati2024llms}. Furthermore, the complexity and scale of disaster environments necessitate an approach that goes beyond simple task execution. Hierarchical Reinforcement Learning (HRL) \cite{HRL_survey} offers a structured method to address this challenge by breaking down complex tasks into more manageable subtasks. HRL’s emphasis on learning at multiple levels of abstraction and operating on a subset of the state space makes it particularly suited for environments with delayed rewards, such as those encountered in SAR operations.

Assuming that the agent is capable of interpreting the information signal into an actionable insight, the retained feedback can promote or discourage behaviour before it is presented \cite{mccallum2023feedback}. However, encoding human input into a shape that an agent understands can be a complicated process. This is where the integration of LLMs with robotic systems marks a significant advancement in the field, bridging the communication gap between humans and machines \cite{wang2024large}. LLMs, especially those tailored with domain-specific knowledge and trained on relevant datasets, show promise and highlight the potential in enhancing decision-making processes in emergency situations \cite{chandra2024exploring}. Recent advancements in integrating LLMs into RL paradigms have shown promise in addressing some of these limitations, as highlighted in \cite{cao2024surveylargelanguagemodelenhanced}. The LLM can act as an information processor, extracting meaningful insights for the agent from natural language, thus enhancing the agent's natural language understanding. However, the potential of combining RL with the nuanced comprehension abilities of LLMs has not been fully exploited. A survey has called for the potential uses of NLP techniques in RL, but the capabilities of LLMs were limited at that time \cite{luketina2019survey}. This synergy \cite{pternea2024rlllmtaxonomytreereviewing} could revolutionise how SAR robots process and act upon human-provided information in real-time.

\subsection{Novelty}
In this paper, the approach considers the agent acting as a first responder in disaster scenarios. Leveraging LLMs to interpret and convert human linguistic inputs into actionable commands enables interaction between SAR robots and non-technical individuals\footnote{those not involved with the agent’s operational software implementation} on-site. By doing so, the proposed system facilitates real-time communication and allows the robot to actively collect, process, and utilise the insights provided by humans, which are crucial in the early stages of disaster response. In addition to that, an HRL framework is adopted to structure the agent's task process \cite{eppe2022intelligent}, effectively addressing the challenges caused by long horizons and sparse rewards. Acting as a first responder in these language-rich environments, the agent uses a hierarchy of decision making to not only solve immediate problems efficiently, as dictated by RL principles, but also to strategically and safely integrate and act upon collected information, improving its learning and operational efficiency.

The main contributions of this paper are summarised as follows: \textbf{(1)} The development of a general, task-oriented hierarchical planning framework for the operation of SAR robots, which includes the use of specialised agents for different subtasks. \textbf{(2)} A novel architecture for human-in-the-loop integration and policy shaping, merging LLMs with HRL to empower SAR robots to interpret and act upon human linguistic inputs in real-time. \textbf{(3)} An extension of the utility of HRL within SAR operations through the proposed method, further enriched by infusing domain-specific knowledge into the LLM using the Retrieval-Augmented Generation (RAG) pipeline \cite{lewis2020retrieval}, enabling the agent to identify, prioritise, and efficiently seek out specific information.

This paper delves into a novel and critical challenge in the realm of emerging crisis response. Crucial information distributed across a crisis scene often remains unexploited, leading to inefficiencies in response strategies and outcomes. By investigating this issue, our work highlights the importance of leveraging this dispersed information within the learning frameworks of SAR operations, ultimately leading to context-sensitive learning. The findings show that descriptive information about the environment (e.g., new hazards or updated victim locations), which may not be immediately available through standard data sources, significantly improves the agent's success on completing the task.

The rest of the paper is organised as follows. Section \ref{methodology} presents an overview of the proposed conceptual architecture. Section \ref{experiments} provides the case study setting followed by experimental evaluation and analysis. In Section \ref{results}, the results of our experiment are presented and discussed and we point directions for future work. Finally, Section \ref{conclusion} concludes our proposed work.

\begin{figure*}[t]
	\centering
    \includegraphics[width=0.99\columnwidth]{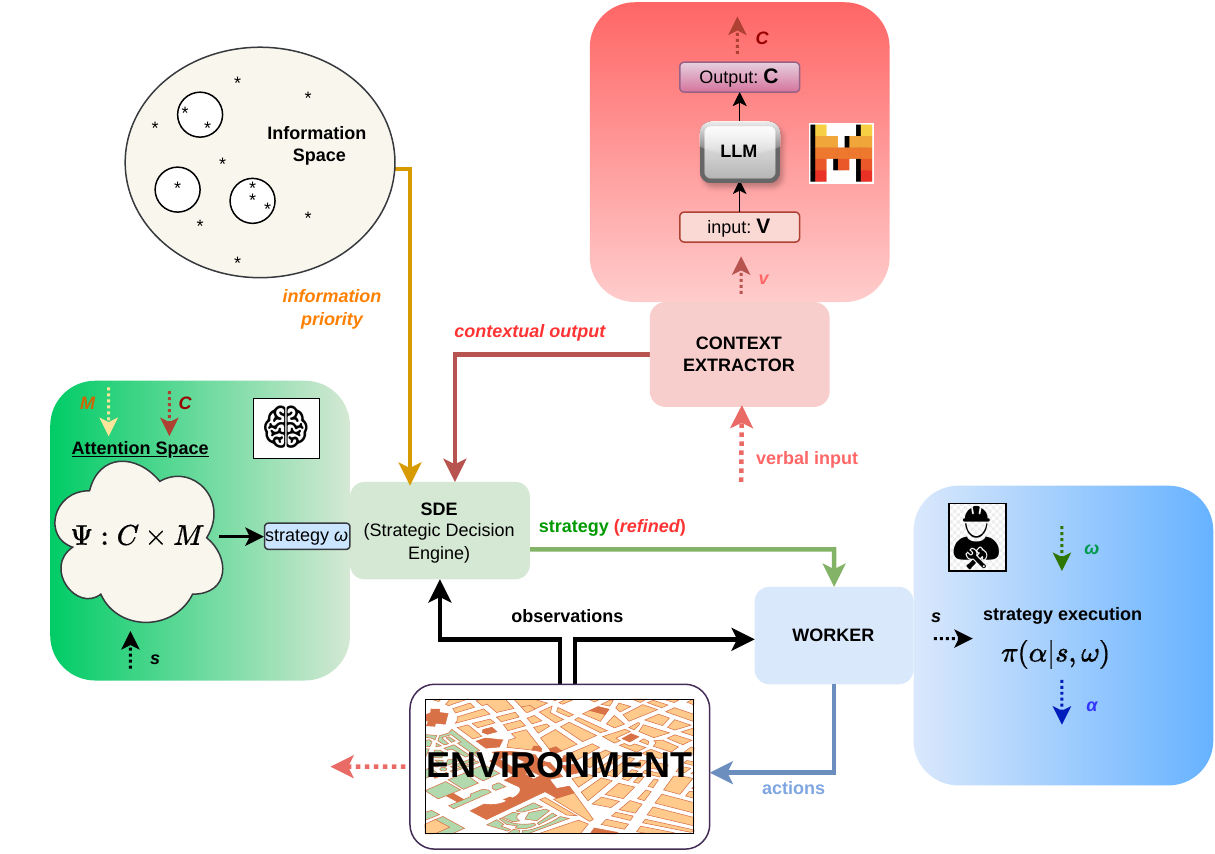}
    \hfill
    \includegraphics[width=0.99\columnwidth]{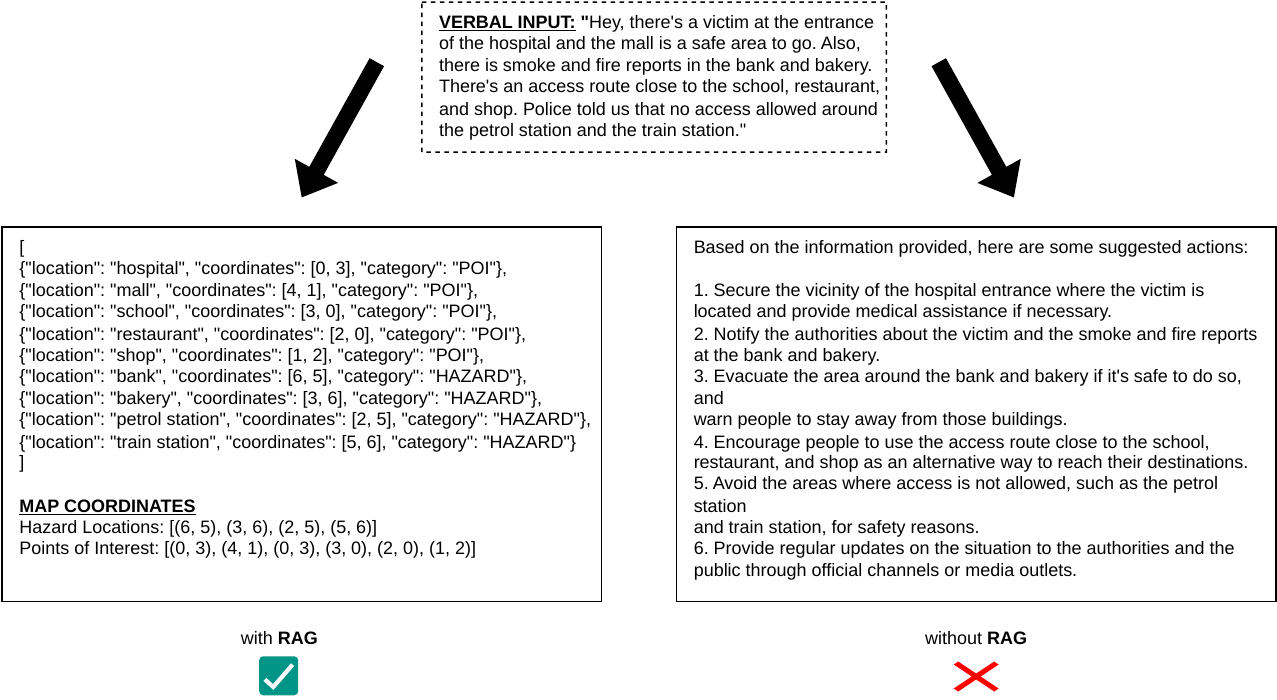}
	\caption{\textit{\textbf{Left:}} The figure illustrates the proposed pipeline within a hierarchical decision-making framework. The \textit{Environment} provides observations $s$ to both the \textit{SDE} and the \textit{Worker} modules. When these observations $s$ contain verbal input $v$, the latter is directed to the \textit{Context Extractor}, which then generates contextual outputs $c$. These outputs $c$, along with observations $s$ from the \textit{Environment} and information priorities $M$ specified by the \textit{Information Space}, are channeled into the SDE. Within the SDE, strategies $\omega$ are refined, taking into account the \textit{Attention Space}. The \textit{Worker} module, informed by these refined strategies $\omega$, executes primitive actions $\alpha$ within the \textit{Environment}. As a result of these interactions, the system continually adjusts and updates its policies, creating a dynamic feedback loop that evolves over time. \textit{\textbf{Right:}} Comparison of outputs from LLMs with and without RAG integration in simulated SAR operations, demonstrating enhanced task-specific detail and notation in outputs.}
	\label{fig:framework}
\end{figure*}

\section{Problem Statement and Modelling}\label{methodology}
\subsection{Conceptual Architecture}
Our aim is to design a novel formulation that incorporates mechanisms to accelerate the learning process of an agent within a HRL framework, integrated with a LLM for processing verbal inputs. The proposed system is illustrated in Fig. \ref{fig:framework} and comprises several key components:

\begin{itemize}
    \item \textbf{Context Extractor:} This module processes verbal inputs communicated to the robot, using a pre-trained LLM to parse and interpret these inputs, and generate a structured contextual representation. The latter, encapsulating key information from the verbal inputs, is then fed to the Strategic Decision Engine (SDE).

    \item \textbf{Information Space:} This set of predefined information types serves as a guide or map for the agent, aligning its actions with the mission's strategic goals. It serves as a reference for the SDE, ensuring that actions remain on track with these objectives.
    
    \item \textbf{Strategic Decision Engine (SDE):} Operating as a manager within the hierarchical framework, the SDE makes strategic decisions based on the environment's state, the context from the Context Extractor, and directives from the Information Space. The decisions guide the agent towards actions aligned with context and mission priorities.
    
    \item \textbf{Attention Space:} Situated within the SDE, this dynamic space influences the agent's decision-making by emphasizing certain aspects of the context and directive information, guiding policy adjustments. It steers the agent towards more context-informed decisions.
    
    \item \textbf{Worker:} This execution module is activated once the SDE selects a strategy, carrying out the corresponding sequence through primitive actions that interact with the environment.

\end{itemize}

\subsection{Modelling}
In HRL, a common approach involves using a hierarchy of decision-makers, such as a manager and workers, where the manager selects which subtask to execute, and each subtask is governed by its own policy. 

We use the formalism of the Markov decision process (MDP), given by a 7-tuple $(S, A, \Omega, \beta_{\omega}, P, R, \gamma)$, where $S$ is the state space, $A$ is the action space, $\Omega$ is the set of strategies each with termination condition $\beta_\omega$, $P$ is the transition function, $R$ is the reward function, $\gamma$ is the discount factor. 

Several elements in the proposed hierarchical framework extend beyond the basic MDP tuple but are crucial for the decision-making process. Specifically, $V$ is a set of verbal inputs $\{v_1, \dots, v_m\}$, where each $v_i$ represents a piece of information related to the task; $C$ denotes contextual details $\{c_1, \dots, c_n\}$, derived from $V$; $M = \{(i_{1}, p_{1}), \dots, (i_{k}, p_{k})\}$ represents the information types and their associated priorities, where $i_{j}$ is the type and $p_{j}$ is its priority; $\pi_\Omega: S\rightarrow\Omega$ is a meta-policy function that maps states to strategies; $\pi_\omega: S\rightarrow A$ is a function that maps states to actions under each strategy $\omega$; $Q(s, \omega)$ evaluates the expected reward of choosing strategy $\omega$ in state $s$; $Q(s, \omega, \alpha)$ evaluates the expected reward of choosing action $\alpha$ under strategy $\omega$ in state $s$; $L: V \rightarrow C$ is a transformation function that processes verbal inputs to generate contextual information; $\Psi$ is the attention space that refines policy functions based on the encoded context and information priorities.

Within this framework, the agent operates on two levels of hierarchy: \textbf{(1)} \textit{High-Level: SDE} The manager decomposes the overall task into smaller, manageable subtasks, sets priorities and the hierarchy for executing these tasks. It selects appropriate strategies for each subtask and monitors and adjusts these strategies based on performance and changes in the environment. \textbf{(2)} \textit{Low-Level: Worker} The worker is responsible for executing the subtasks assigned by the manager. This involves directly interacting with the environment through specific actions determined by sub-policies $\pi_{\omega}$. Both layers can either be learned over time through experience or follow deterministic rules.

The objective is to find the optimal policies $\pi_\Omega$ and $\pi_\omega$ that maximise the expected reward over time, while also considering the influence of the attention space $\Psi$ in dynamically refining and guiding the policy selection process based on context and information priorities:

\begin{equation}
J(\pi_\Omega, \pi_\omega) = \mathbb{E} \left[\sum_{t=0}^{T} \gamma^{t} R(s_t, a_t) \mid \pi^{\Psi}_\Omega, \pi^{\Psi}_\omega \right]
\end{equation}

where $\gamma$ is the discount factor, and $T$ the time horizon. The action $\alpha_t$ at time $t$ is determined by the sub-policy $\pi_\omega$ if the policy $\omega$ is active, and $s_t$ is the state of the environment at time $t$.

\begin{figure*}[t]
    \centering    \includegraphics[width=0.3\textwidth,height=0.22\textheight]{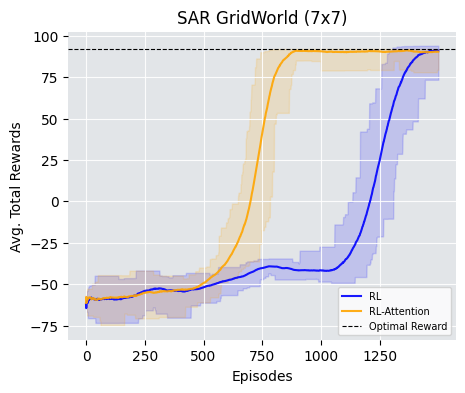}
    \hfill    \includegraphics[width=0.3\textwidth,height=0.22\textheight]{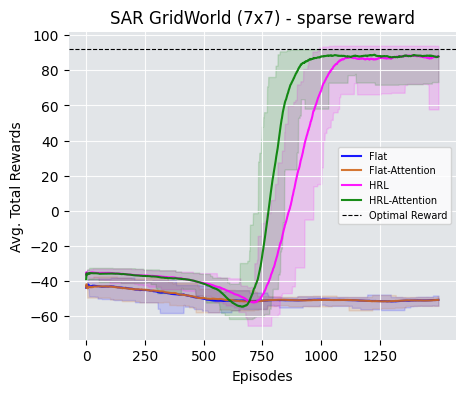}
    \hfill    \includegraphics[width=0.3\textwidth,height=0.21\textheight]{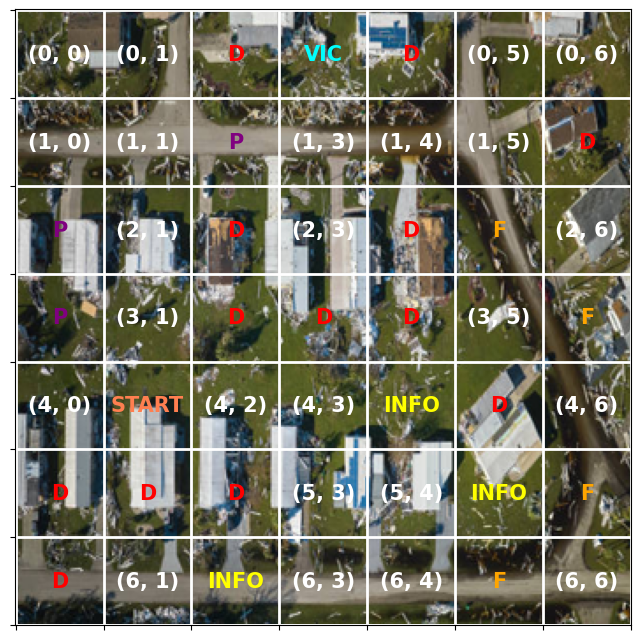}
    \caption{Comparative Analysis of Learning Agents in SAR Scenarios. \textit{\textbf{Left:}} Performance of flat RL agents, with and without attention guidance, receiving intrinsic rewards. \textit{\textbf{Middle:}} Comparison of hierarchical (HRL) and flat RL agents, with and without attention guidance under sparse reward - reward is given upon successful task completion - conditions. \textit{\textbf{Right:}}  SAR environment configuration showing information locations marked as `INFO', obstacles `D', victim locations `VIC', hazards `F', and points of interest `P'.}
    \label{fig:rl_rewards}
\end{figure*}

\section{Experiments} \label{experiments}
\subsection{Simulated Environment Setup}
To evaluate the efficiency of our proposed system integrating LLM with a hierarchical framework, we design a SAR simulated environment with discrete state and action space. Specifically, in the designed 2D experimental setup (see Fig. \ref{fig:rl_rewards}), an agent is tasked with operating in a disaster-stricken area, aiming to rescue victims while avoiding obstacles. The experiment simulates a complex scenario, where the agent must not only locate and save victims but also optimise the collection and processing of crucial information within the environment. As the agent navigates through designated information locations, they must learn to gather data prioritised by a predefined information space. This data includes updates on victim locations, safe routes, and potential hazards, mimicking real-world SAR operations where such information is crucial for effective decision-making. Upon receiving verbal information, the hazards and points of interest are revealed to the agent and are marked as `F' and `P', respectively. The agent transforms this input into contextual representation, allowing it to dynamically adapt and enhance its decision-making process through an attention space. This space refines the policies within the hierarchical framework, steering the agent's behavior towards more context-informed decisions by exploiting the feedback. While we restrict our analysis to a simplified toy example, allowing for complete control over all variables and extensive experimentation across various configurations, this simulation is intended to replicate the complex nature of actual disaster situations. The development of the environment largely adheres to the structure and conventions of a custom environment in the OpenAI Gym framework \cite{brockman2016openai}. The code is available at \url{https://github.com/dimipan/HRL-LLM}.


\begin{table}[h]
\caption{Performance metrics of different models}
    \centering
    \small
    \begin{tabular}{lccc}
        \toprule
        \textbf{Model} & \textbf{Collisions} & \textbf{Steps} & \textbf{Avg. Reward} \\
        \midrule
        Flat (no sparse)      & 3 & 24 & -19.8 \\
        Flat-Att (no sparse)  & 0 & 26 & 26.5 \\
        HRL (sparse)          & 3 & 24 & 14.12 \\
        HRL-Att (sparse)      & 0 & 26 & 22.8 \\
        \bottomrule
    \end{tabular}
    \label{tab:model_performance}
\end{table}

\subsection{Implementation Details}
Here, we elaborate on the functionality of key elements and demonstrate the practical application of our proposed system through a case study. All components of the system run on a single machine equipped with an RTX A2000 GPU. Our system employs the Mistral 7B \cite{jiang2023mistral} pre-trained LLM, integrated within a local RAG pipeline. Embedding models and LLMs are hosted using Ollama, with a local Chroma instance as the vector store, and Langchain orchestrates all processes. The model was selected for its suitability in real-time applications, offering quick responses and outperforming even models with larger parameter counts. Additionally, we infuse a JSON object containing structured data with sets of critical keywords and mapped locations, complete with precise coordinates. The information space categorises essential SAR information \cite{katsadouros2022introducing} into three types: victim details (X), navigation routes (Y), and environmental hazards (Z), with the requirement that this sequence must be respected during the collection process. The SDE dynamically selects strategies based on conditions reflected in the agent's state representation. The latter includes the agent's position on the grid, the status of each type of information—indicated in a binary format to show whether it has been collected or not—and an additional indicator that specifies whether a victim has been saved or not. Specifically, strategy $\pi_{EXP}$ is selected when the agent needs to navigate to information points or the final location. Strategy $\pi_{COL}$ is chosen when the agent reaches information points to gather data. Strategy $\pi_{OPE}$ is selected when the agent must perform critical triage tasks. 
The worker interacts with the environment by executing the strategies decided by the SDE through primitive actions. In particular, these actions are abstracted into different types depending on the current strategy. Within strategy \textit{`EXPLORE'} movement actions \textit{\{`up', `down', `left', `right'\}} are enabled. When strategy \textit{`COLLECT'} is active a set of actions \textit{\{`A', `B', `C', `X', `Y', `Z'\}} reflects the agent's ability to get information. Finally, within strategy \textit{`OPERATE'} the agent can choose from \textit{\{`save', `use', `remove', `carry'\}}. Both strategies and primitive actions can be refined by the attention space via policy shaping, allowing the agent to prioritise certain aspects according to the context extracted from human feedback. The execution of each worker associated with a strategy is trained and learned through RL. The training process entails running the system across 1500 episodes for a total of 50 runs, utilising the Q-learning algorithm. Throughout these episodes, the performance of the system is evaluated by averaging the rewards obtained in each run. A decaying $\epsilon$-greedy action selection method is employed with an initial $\epsilon$ value 1.0 and linearly decaying to a minimum of 0.01 at a decay rate of 2. The discount factor $\gamma$ and learning rate $\alpha$ are set at $0.998$ and $0.1$, respectively. When the agent exploits the collected information and acts based on the attention space, the $\epsilon$ decay is steeper.

\subsection{Hypotheses}
In our experimental evaluation, we study the following hypotheses: \textbf{(1)} The use of an LLM infused with domain-specific knowledge through the RAG pipeline produces context-informed outputs that are more relevant and accurate in SAR scenarios than those generated without RAG \cite{gao2023retrieval}. \textbf{(2)} The integration of the attention space into a flat RL agent accelerates the learning process, resulting in better and faster convergence. \textbf{(3)} The hierarchical setup is particularly effective in sparse reward environments, where rewards are only granted upon task completion, outperforming flat RL setups. \textbf{(4)} The integration of the attention space into the hierarchical further improves the agent's performance. \textbf{(5)} Utilising the attention space in the decision making leads to a reduction in encounters with dynamic obstacles, highlighting the value of real-time, contextually aware feedback in operational settings.

\section{Results \& Discussion}\label{results}
We compare the performance of agents in both flat and hierarchical structures, with and without attention guidance. The experiments were tailored to test specific hypotheses mentioned above related to the efficacy of incorporating domain-specific knowledge and attention mechanisms into  learning agents.

\subsection{Hypothesis 1: Domain-Knowledge Infused LLMs}
The first hypothesis suggests that in simulated SAR operations LLMs would lack the capability to produce context-informed outputs properly aligned with the demands of the task without the RAG integration. Fig. \ref{fig:framework} supports this hypothesis and shows the practical enhancements in output accuracy and detail achieved through RAG integration.    

\subsection{Hypothesis 2: RL with Attention Space} 
For the second hypothesis, we expect that integrating the LLM into a flat RL agent with an attention space would accelerate the learning process. The results from Fig. \ref{fig:rl_rewards} indicate that agents guided by the attention space not only perform better in terms of reward obtained across episodes but also show improved learning speed. This improvement is particularly evident in scenarios where agents receives ongoing intrinsic rewards throughout the episode.

\subsection{Hypothesis 3 \& 4: HRL (with Attention Space) in Sparse Reward Environment}
The third and fourth hypotheses examine the effectiveness of hierarchical task formulation in sparse reward environments, both with and without the integration of LLM and attention space. Hierarchical agents consistently outperform flat agents, which fail to learn any effective policy in environments where rewards are only dispensed upon the completion of tasks. This highlights the effectiveness of hierarchical structures in managing sparse reward environments. The addition of LLM and attention space further boosts the performance of the hierarchical agents as results indicate.

\subsection{Hypothesis 5: Safe Navigation}
Lastly, the fifth hypothesis addresses the system's potential to reduce encounters with dynamic obstacles, as detailed in Table \ref{tab:model_performance}. The number of collisions and steps taken refer to the learned policy, while the average reward is over the entire learning curve. The results show that the use of LLM and attention space significantly reduces the frequency of these encounters. It is worth pointing out that for the flat agents, we only consider the intrinsic reward setup where they successfully converge, thus disregarding the sparse reward configuration. Both flat and hierarchical agents without attention mechanisms complete the task with fewer steps (24 steps on average), but frequently collide with dynamic obstacles that become apparent after verbal input is communicated (3 collisions on average). In contrast, flat and hierarchical agents utilising attention mechanisms take slightly more steps to complete the task (26 steps on average) but successfully avoid collisions (0 collisions on average), demonstrating their ability to effectively use real-time, context-aware feedback to adapt their policies.

The results from these experiments highlight the potential of integrating advanced LLMs and attention mechanisms into RL systems, particularly in challenging and dynamic environments like SAR operations. Context-aware decision-making capabilities facilitated by these refinements not only improve the performance in terms of task-specific metrics but also enhance the adaptability and safety of agents operating in real-world conditions. The demonstrated effectiveness of hierarchical structures in managing complex and sparse reward scenarios suggests a promising avenue for developing more robust systems for HRI. 

While our experiments provide a validation of our hypotheses, there are several limitations to address. In particular, when validating these systems in continuous domains employing deep RL techniques, policy shaping is intricate and less intuitive. Therefore, it may be advantageous to prefer action space bounding or shaping over policy shaping, as discussed here. This approach allows for a more straightforward implementation, since actions can be intuitively bounded based on well-defined physical constraints or safety requirements.

Another challenge emerges with the use of language. Real-world deployment in such scenarios often entails interacting with non-standardised, potentially unreliable linguistic inputs from humans. This can mislead the system and make the extraction of actionable information complicated. A solution to that could be the infusion of additional documents, expert knowledge, and detailed contextual data into the LLM.

Lastly, the computational cost of employing more advanced LLMs with larger numbers of parameters must also be acknowledged. While these models may offer superior reasoning performance and better natural language understanding, their computational and resource demands are significantly higher.

\section{CONCLUSIONS}\label{conclusion}
This paper investigates the integration of LLMs and hierarchical learning within the context of SAR operations, demonstrating how the sophisticated interplay between advanced computational tools and human input can revolutionise SAR autonomous systems. Our approach, specifically, addresses the critical need for SAR robots to adapt rapidly and make context-informed decisions in dynamic disaster scenarios. The system’s ability to prioritise predefined types of information and adjust its strategy based on immediate feedback marks a significant advancement over traditional methods, achieving a balance between rapid response capabilities and strategic information gathering. Our work highlights the untapped potential of leveraging advanced language and reasoning capabilities of LLMs to convert verbal inputs into actionable insights. By optimising learning and decision-making processes and introducing an attention space that refines strategy based on context, our paper opens new avenues for more context-sensitive and efficient robotic responses in high-stakes environments, potentially setting new standards for future implementations.




\addtolength{\textheight}{-12cm}   




\bibliography{thebibliography}

\begin{thebibliography}{10}
\providecommand{\url}[1]{#1}
\csname url@samestyle\endcsname
\providecommand{\newblock}{\relax}
\providecommand{\bibinfo}[2]{#2}
\providecommand{\BIBentrySTDinterwordspacing}{\spaceskip=0pt\relax}
\providecommand{\BIBentryALTinterwordstretchfactor}{4}
\providecommand{\BIBentryALTinterwordspacing}{\spaceskip=\fontdimen2\font plus
\BIBentryALTinterwordstretchfactor\fontdimen3\font minus \fontdimen4\font\relax}
\providecommand{\BIBforeignlanguage}[2]{{%
\expandafter\ifx\csname l@#1\endcsname\relax
\typeout{** WARNING: IEEEtran.bst: No hyphenation pattern has been}%
\typeout{** loaded for the language `#1'. Using the pattern for}%
\typeout{** the default language instead.}%
\else
\language=\csname l@#1\endcsname
\fi
#2}}
\providecommand{\BIBdecl}{\relax}
\BIBdecl

\bibitem{SOORI202354}
\BIBentryALTinterwordspacing
M.~Soori, B.~Arezoo, and R.~Dastres, ``Artificial intelligence, machine learning and deep learning in advanced robotics, a review,'' \emph{Cognitive Robotics}, vol.~3, pp. 54--70, 2023. [Online]. Available: \url{https://www.sciencedirect.com/science/article/pii/S2667241323000113}
\BIBentrySTDinterwordspacing

\bibitem{el2024advanced}
M.~El~Debeiki, S.~Al-Rubaye, A.~Perrusqu{\'\i}a, C.~Conrad, and J.~A. Flores-Campos, ``An advanced path planning and uav relay system: Enhancing connectivity in rural environments,'' \emph{Future Internet}, vol.~16, no.~3, p.~89, 2024.

\bibitem{bridge}
P.~Arya, W.~Guo, S.~K. Ahmed, and L.~Irwanda, ``Deep learning for bridge load capacity estimation in post-disaster and -conflict zones,'' \emph{Royal Society Open Science}, vol.~6, 2019.

\bibitem{damage}
L.~Lu and W.~Guo, ``Automatic quantification of settlement damage using deep learning of satellite images,'' in \emph{2021 IEEE International Smart Cities Conference (ISC2)}, 2021, pp. 1--6.

\bibitem{liu2013robotic}
Y.~Liu and G.~Nejat, ``Robotic urban search and rescue: A survey from the control perspective,'' \emph{Journal of Intelligent \& Robotic Systems}, vol.~72, pp. 147--165, 2013.

\bibitem{xi2023rise}
Z.~X. et~al., ``The rise and potential of large language model based agents: A survey,'' 2023.

\bibitem{zhang2019leveraging}
R.~Zhang, F.~Torabi, L.~Guan, D.~H. Ballard, and P.~Stone, ``Leveraging human guidance for deep reinforcement learning tasks,'' \emph{arXiv preprint arXiv:1909.09906}, 2019.

\bibitem{human-SAR}
R.~Krzysiak and S.~Butail, ``Information-based control of robots in search-and-rescue missions with human prior knowledge,'' \emph{IEEE Transactions on Human-Machine Systems}, vol.~52, no.~1, pp. 52--63, 2022.

\bibitem{bignold2022human}
A.~Bignold, F.~Cruz, R.~Dazeley, P.~Vamplew, and C.~Foale, ``Human engagement providing evaluative and informative advice for interactive reinforcement learning,'' \emph{Neural Computing and Applications}, pp. 1--16, 2022.

\bibitem{guo2022survey}
Z.~Guo, C.~Yao, Y.~Feng, and Y.~Xu, ``Survey of reinforcement learning based on human prior knowledge,'' \emph{Journal of Uncertain Systems}, vol.~15, no.~01, p. 2230001, 2022.

\bibitem{phone}
A.~Albanese, V.~Sciancalepore, and X.~Costa-Pérez, ``Sardo: An automated search-and-rescue drone-based solution for victims localization,'' \emph{IEEE Transactions on Mobile Computing}, vol.~21, no.~9, pp. 3312--3325, 2022.

\bibitem{10.1371/journal.pone.0273696}
\BIBentryALTinterwordspacing
L.~Lawrie, K.~Gillies, E.~Duncan, L.~Davies, D.~Beard, and M.~K. Campbell, ``Barriers and enablers to the effective implementation of robotic assisted surgery,'' \emph{PLOS ONE}, vol.~17, no.~8, pp. 1--21, 08 2022. [Online]. Available: \url{https://doi.org/10.1371/journal.pone.0273696}
\BIBentrySTDinterwordspacing

\bibitem{Kozlowski2006Teams}
\BIBentryALTinterwordspacing
S.~W. Kozlowski and D.~R. Ilgen, ``Enhancing the effectiveness of work groups and teams,'' \emph{Psychological Science in the Public Interest}, vol.~7, no.~3, pp. 77--124, 2006, pMID: 26158912. [Online]. Available: \url{https://doi.org/10.1111/j.1529-1006.2006.00030.x}
\BIBentrySTDinterwordspacing

\bibitem{waring2020role}
S.~Waring, L.~Alison, N.~Shortland, and M.~Humann, ``The role of information sharing on decision delay during multiteam disaster response,'' \emph{Cognition, Technology \& Work}, vol.~22, pp. 263--279, 2020.

\bibitem{LLMsEvaluation}
\BIBentryALTinterwordspacing
Y.~Chang, X.~Wang, J.~Wang, Y.~Wu, L.~Yang, K.~Zhu, H.~Chen, X.~Yi, C.~Wang, Y.~Wang, W.~Ye, Y.~Zhang, Y.~Chang, P.~S. Yu, Q.~Yang, and X.~Xie, ``A survey on evaluation of large language models,'' \emph{ACM Trans. Intell. Syst. Technol.}, vol.~15, no.~3, mar 2024. [Online]. Available: \url{https://doi.org/10.1145/3641289}
\BIBentrySTDinterwordspacing

\bibitem{kambhampati2024llms}
S.~Kambhampati, K.~Valmeekam, L.~Guan, K.~Stechly, M.~Verma, S.~Bhambri, L.~Saldyt, and A.~Murthy, ``Llms can't plan, but can help planning in llm-modulo frameworks,'' 2024.

\bibitem{HRL_survey}
\BIBentryALTinterwordspacing
S.~Pateria, B.~Subagdja, A.-h. Tan, and C.~Quek, ``Hierarchical reinforcement learning: A comprehensive survey,'' \emph{ACM Comput. Surv.}, vol.~54, no.~5, jun 2021. [Online]. Available: \url{https://doi.org/10.1145/3453160}
\BIBentrySTDinterwordspacing

\bibitem{mccallum2023feedback}
S.~McCallum, M.~Taylor-Davies, S.~Albrecht, and A.~Suglia, ``Is feedback all you need? leveraging natural language feedback in goal-conditioned rl,'' in \emph{NeurIPS 2023 Workshop on Goal-Conditioned Reinforcement Learning}, 2023.

\bibitem{wang2024large}
J.~Wang, Z.~Wu, Y.~Li, H.~Jiang, P.~Shu, E.~Shi, H.~Hu, C.~Ma, Y.~Liu, X.~Wang \emph{et~al.}, ``Large language models for robotics: Opportunities, challenges, and perspectives,'' \emph{arXiv preprint arXiv:2401.04334}, 2024.

\bibitem{chandra2024exploring}
A.~Chandra and A.~Chakraborty, ``Exploring the role of large language models in radiation emergency response,'' \emph{Journal of Radiological Protection}, 2024.

\bibitem{cao2024surveylargelanguagemodelenhanced}
\BIBentryALTinterwordspacing
Y.~Cao, H.~Zhao, Y.~Cheng, T.~Shu, G.~Liu, G.~Liang, J.~Zhao, and Y.~Li, ``Survey on large language model-enhanced reinforcement learning: Concept, taxonomy, and methods,'' 2024. [Online]. Available: \url{https://arxiv.org/abs/2404.00282}
\BIBentrySTDinterwordspacing

\bibitem{luketina2019survey}
J.~Luketina, N.~Nardelli, G.~Farquhar, J.~Foerster, J.~Andreas, E.~Grefenstette, S.~Whiteson, and T.~Rockt{\"a}schel, ``A survey of reinforcement learning informed by natural language,'' \emph{arXiv preprint arXiv:1906.03926}, 2019.

\bibitem{pternea2024rlllmtaxonomytreereviewing}
\BIBentryALTinterwordspacing
M.~Pternea, P.~Singh, A.~Chakraborty, Y.~Oruganti, M.~Milletari, S.~Bapat, and K.~Jiang, ``The rl/llm taxonomy tree: Reviewing synergies between reinforcement learning and large language models,'' 2024. [Online]. Available: \url{https://arxiv.org/abs/2402.01874}
\BIBentrySTDinterwordspacing

\bibitem{eppe2022intelligent}
M.~Eppe, C.~Gumbsch, M.~Kerzel, P.~D. Nguyen, M.~V. Butz, and S.~Wermter, ``Intelligent problem-solving as integrated hierarchical reinforcement learning,'' \emph{Nature Machine Intelligence}, vol.~4, no.~1, pp. 11--20, 2022.

\bibitem{lewis2020retrieval}
P.~Lewis, E.~Perez, A.~Piktus, F.~Petroni, V.~Karpukhin, N.~Goyal, H.~K{\"u}ttler, M.~Lewis, W.-t. Yih, T.~Rockt{\"a}schel \emph{et~al.}, ``Retrieval-augmented generation for knowledge-intensive nlp tasks,'' \emph{Advances in Neural Information Processing Systems}, vol.~33, pp. 9459--9474, 2020.

\bibitem{brockman2016openai}
G.~Brockman, V.~Cheung, L.~Pettersson, J.~Schneider, J.~Schulman, J.~Tang, and W.~Zaremba, ``Openai gym,'' \emph{arXiv preprint arXiv:1606.01540}, 2016.

\bibitem{jiang2023mistral}
A.~Q. Jiang, A.~Sablayrolles, A.~Mensch, C.~Bamford, D.~S. Chaplot, D.~d.~l. Casas, F.~Bressand, G.~Lengyel, G.~Lample, L.~Saulnier \emph{et~al.}, ``Mistral 7b,'' \emph{arXiv preprint arXiv:2310.06825}, 2023.

\bibitem{katsadouros2022introducing}
E.~Katsadouros, D.~G. Kogias, C.~Z. Patrikakis, G.~Giunta, A.~Dimou, and P.~Daras, ``Introducing the architecture of faster: A digital ecosystem for first responder teams,'' \emph{Information}, vol.~13, no.~3, p. 115, 2022.

\bibitem{gao2023retrieval}
Y.~Gao, Y.~Xiong, X.~Gao, K.~Jia, J.~Pan, Y.~Bi, Y.~Dai, J.~Sun, and H.~Wang, ``Retrieval-augmented generation for large language models: A survey,'' \emph{arXiv preprint arXiv:2312.10997}, 2023.

\end{thebibliography}
\bibliographystyle{IEEEtran}

\end{document}